\documentclass[10pt,twocolumn,letterpaper]{article}

\usepackage{cvpr}
\usepackage{times}
\usepackage{epsfig}
\usepackage{graphicx}
\usepackage{amsmath}
\usepackage{amssymb}

\usepackage[lined,boxed,commentsnumbered]{algorithm2e}
\usepackage{float}
\newfloat{myalgo}{tbhp}{mya}
{\begin{myalgo}[#1]
\centering
\begin{minipage}{1.\linewidth}
\begin{algorithm}[H]}%
{\end{algorithm}
\end{minipage}
\end{myalgo}}

\usepackage{enumitem}
\usepackage{tabularx}
\usepackage{multirow}
\usepackage{makecell}

\usepackage[pagebackref=true,breaklinks=true,letterpaper=true,colorlinks,bookmarks=false]{hyperref}

\cvprfinalcopy 

\def\cvprPaperID{3256} 

\ifcvprfinal\fi
\begin{document}


\title{AOGNets: Compositional Grammatical Architectures for Deep Learning}

\author{Xilai Li$^{\dagger}$, Xi Song$^\mathsection$ and Tianfu Wu$^{\dagger,\ddagger}$\thanks{T. Wu is the corresponding author. $^\mathsection$X. Song is an independent researcher. The code and models are available at \url{https://github.com/iVMCL/AOGNets}}\\
Department of ECE$^\dagger$ and the Visual Narrative Initiative$^\ddagger$, 
North Carolina State University\\
{\tt\small \{xli47, tianfu\_wu\}@ncsu.edu, xsong.lhi@gmail.com}
}

\maketitle

\begin{abstract}
    Neural architectures are the foundation for improving performance of deep neural networks (DNNs). This paper presents deep compositional grammatical architectures which harness the best of two worlds: grammar models and DNNs. The proposed architectures integrate compositionality and reconfigurability of the former and the capability of learning rich features of the latter in a principled way. 
    We utilize AND-OR Grammar (AOG)~\cite{DisAOT-CVPR,Zhu_Grammar,Yuille_AndOr} as network generator in this paper and call the resulting networks \textbf{AOGNets}. An AOGNet consists of a number of stages each of which is composed of a number of AOG building blocks. An AOG building block splits its input feature map into $N$ groups along feature channels and then treat it as a sentence of $N$ words. It then jointly realizes a phrase structure grammar and a dependency grammar in bottom-up parsing the ``sentence" for better feature exploration and reuse. 
    It provides a unified framework for the best practices developed in state-of-the-art DNNs. 
    In experiments, AOGNet is tested in the CIFAR-10, CIFAR-100  and ImageNet-1K classification benchmark  and the MS-COCO object detection and segmentation benchmark. In CIFAR-10, CIFAR-100 and ImageNet-1K, AOGNet obtains better performance than ResNet~\cite{ResidualNet} and most of its variants, ResNeXt~\cite{ResNeXt} and its attention based variants such as SENet~\cite{SENet}, DenseNet~\cite{DenseNet} and DualPathNet~\cite{DPN}. AOGNet also obtains the best model interpretability score using network dissection~\cite{netdissect2017}. AOGNet further shows better potential in adversarial defense. In MS-COCO, AOGNet obtains better performance than the ResNet and ResNeXt backbones in Mask R-CNN~\cite{MaskRCNN}. 
\end{abstract}

\section{Introduction}
\vspace{-1mm}
\subsection{Motivation and Objective} 
\begin{figure}
    	\centering
    	\includegraphics[width = 0.5\textwidth]{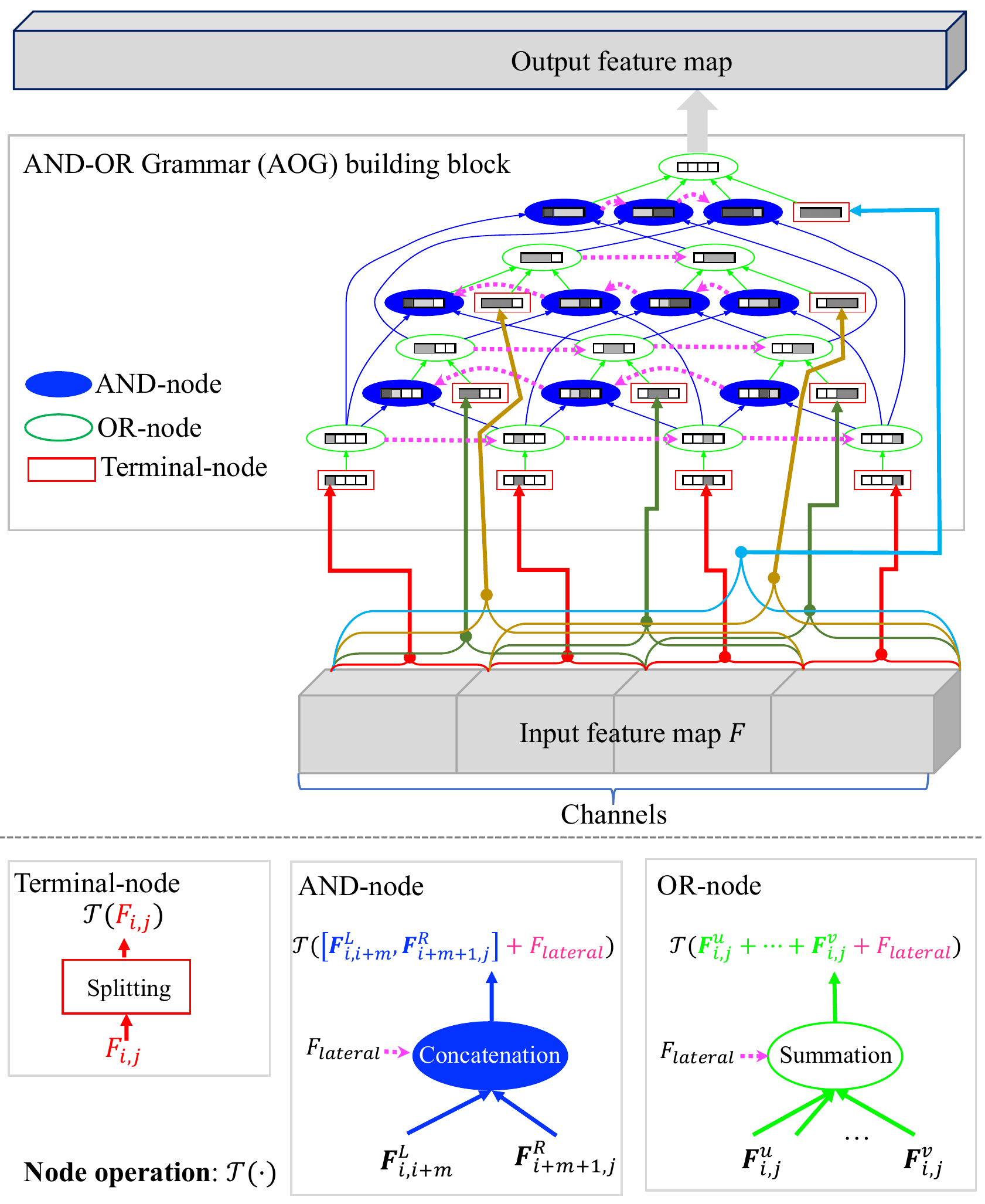}
    	\caption{\small{Illustration of our AOG building block for grammar-guided network generator. The resulting networks, AOGNets obtain $80.18$\% top-1 accuracy with 40.3M parameters in ImageNet, significantly outperforming ResNet-152 ($77.0$\%, 60.2M), ResNeXt-101 ($79.6$\%, 83.9M), DenseNet-Cosine-264 ($79.6$\%, $\sim$73M) and DualPathNet-98 ($79.85$\%, 61.6M).    
    	See text for details. (Best viewed in color)} 
    	}
    	\label{fig:AOG-block} 
\end{figure}
Recently, deep neural networks (DNNs)~\cite{LeCunCNN,AlexNet} have improved prediction accuracy significantly in many vision tasks, and have obtained superhuman performance in image classification tasks~\cite{ResidualNet,InceptionNet,DenseNet,DPN}.
Much of these progress are achieved mainly through engineering network architectures which jointly address two issues: increasing representational power by going either deeper or wider, and maintaining the feasibility of optimization using back-propagation with stochastic gradient descent (i.e., the  vanishing and/or exploding gradient problems). 
The dramatic success does not necessarily speak to its sufficiency given the lack of theoretical underpinnings of DNNs at present~\cite{BoundsOfDL}. 
Different methodologies are worth exploring to enlarge the scope of neural architectures for seeking better DNNs. For example, 
Hinton recently pointed out a crucial drawback of  current convolutional neural networks:
according to recent neuroscientific research, these artificial networks do not contain enough levels of structure~\cite{Hinton,capsules}. In this paper, we are interested in \textbf{grammar-guided network generators} (Fig.~\ref{fig:AOG-block}).

\begin{figure}
    	\centering
    	\includegraphics[width = 0.5\textwidth]{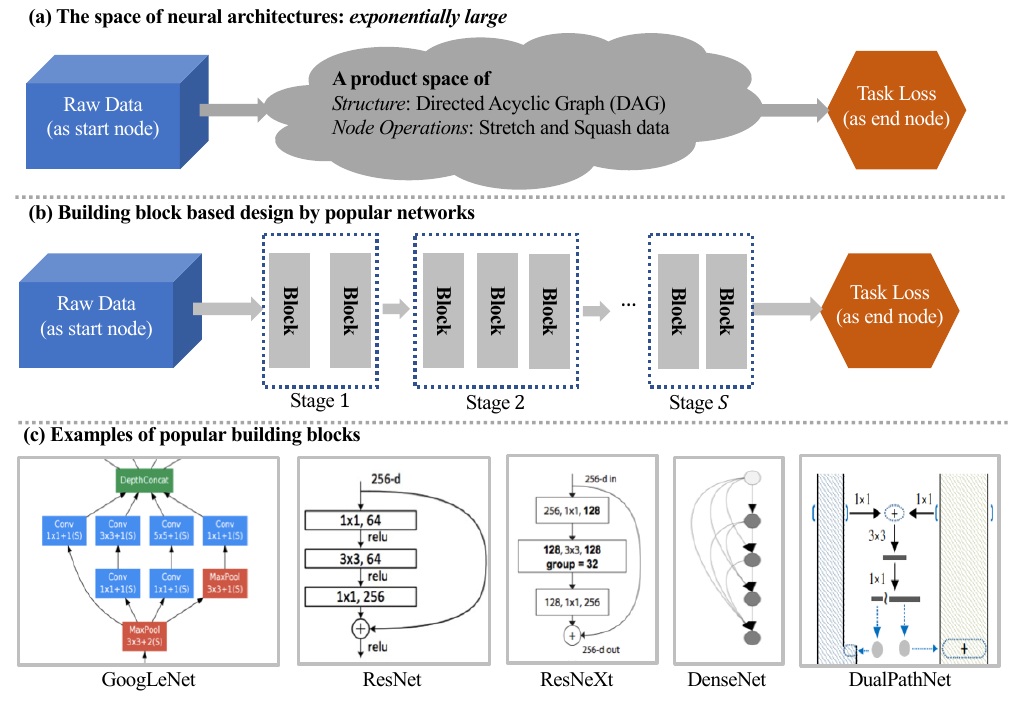}
    	\caption{\small{Illustration of (a) the space of neural architectures, (b) the building block based design, and (c) examples of popular building blocks in GoogLeNet~\cite{InceptionNet}, ResNet~\cite{ResidualNet}, ResNeXt~\cite{ResNeXt}, DenseNet~\cite{DenseNet} and DualPathNets~\cite{DPN}.
    	See text for details.} 
    	}
    	\label{fig:existing-blocks} 
\end{figure}

Neural architecture design and search can be posed as a combinatorial search problem in a product space comprising two sub-spaces (Fig.~\ref{fig:existing-blocks} (a)): 

\begin{itemize} 
    \item The structure space which consists of all directed acyclic graphs (DAGs) with the start node representing input raw data and the end node representing task loss functions. DAGs are entailed for feasible computation.
    \item The node operation space which consists of all possible transformation functions for implementing nodes in a DAG, such as Convolution+BatchNorm~\cite{BatchNorm}+ReLU~\cite{AlexNet} and its bottleneck implementation~\cite{ResidualNet} with different kernel sizes and different numbers of feature channels. 
\end{itemize}

The structure space is almost unbounded, and the node operation space for a given structure is also combinatorial. Neural architecture design and search is a challenging problem due to the exponentially large space and the highly non-convex non-linear objective function to be optimized in the search. As illustrated in Fig.~\ref{fig:existing-blocks} (b), to mitigate the difficulty, neural architecture design and search have been simplified to design or search a building block structure. Then, a DNN consists of a predefined number of stages each of which has a small number of building blocks. This stage-wise building-block based design is also supported by the theoretical study in~\cite{BoundsOfDL} under some assumptions. Fig.~\ref{fig:existing-blocks} (c) shows examples of some popular building blocks with different structures. Two questions arise naturally: 
\begin{itemize} 
\item Can we unify the best practices used by the popular building blocks in a simple and elegant framework? More importantly, can we generate building blocks and thus networks in a principled way to effectively unfold the space (Fig.~\ref{fig:existing-blocks} (a)) ?  (If doable)
\item Will the unified building block/network generator improve performance on accuracy, model interpretability and adversarial robustness without increasing model complexities and computational costs? If yes, the potential impacts shall be broad and deep for representation learning in numerous practical applications. 
\end{itemize}

To address the above questions, we first need to understand the underlying wisdom in designing better network architectures: It usually lies in finding network structures which can support flexible and diverse information flows for exploring new features, reusing existing features in previous layers and back-propagating learning signals (e.g., gradients). Then, what are the key principles that we need to exploit and formulate such that we can effectively and efficiently unfold the structure space in Fig.~\ref{fig:existing-blocks} (a) in a way better than existing networks?
\textit{Compositionality, reconfigurability and lateral connectivity} are well-known principles in cognitive science, neuroscience and pattern theory~\cite{Geman_CompositionSystems,Mumford_PT,Grenander_PT,RCN,ProbabilisticProgramInduction,RCN}. They are fundamental for the remarkable capabilities possessed by humans, of learning rich knowledge and adapting to different environments, especially in vision and language. They have not been, however, fully and explicitly integrated in DNNs.

\begin{figure*} [t]
    	\centering
    	\includegraphics[width = 1.0\textwidth]{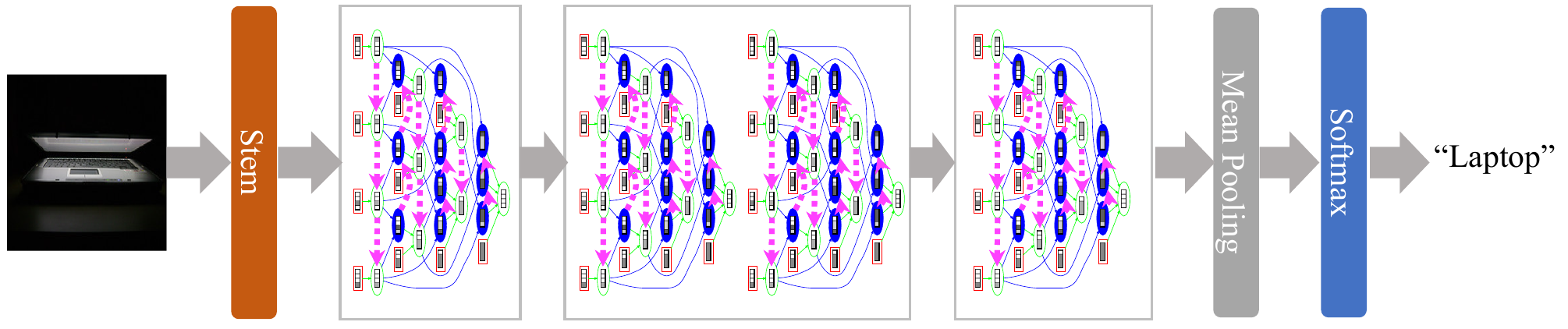}
    	\caption{\small{Illustration of a 3-stage AOGNet with 1 AOG building bock in the 1st and 3rd stage, and 2 AOG building blocks in the 2nd stage. Note that different stages can use different AOG building blocks. We show the same one for simplicity. The stem can be either a vanilla convolution or convolution+MaxPooling.  
    	(Best viewed in color) }
    	}
    	\label{fig:model} 
\end{figure*}

In this paper, we presents \textbf{compositional grammatical architectures} that realize compositionality, reconfigurability and lateral connectivity for building block design in a principled way. We utilize AND-OR Grammars (AOG)~\cite{DisAOT-CVPR,Zhu_Grammar,Yuille_AndOr} and propose AOG building blocks that unify the best practices developed in existing popular building blocks. Our method deeply integrates hierarchical and compositional grammars and DNNs for harnessing the best of both worlds in deep representation learning. 

Why grammars? 
Grammar models are well known in both natural language processing and computer vision. Image grammar~\cite{Zhu_Grammar,Pff_Grammar,Yuille_AndOr,Geman_CompositionSystems} was one of the dominant methods in computer vision before the recent resurgence in popularity of deep neural networks. With the recent resurgence, one fundamental puzzle arises that grammar models with more explicitly compositional structures and better analytic and theoretical potential, often perform worse than their neural network counterparts. 
As David Mumford pointed out, ``\textit{Grammar in language is merely a recent extension of much older grammars that are built into the brains of all intelligent animals to analyze sensory input, to structure their actions and even formulate their thoughts.}"~\cite{Mumford}. Our proposed AOG building block is highly expressive for analyzing sensory input and bridges the performance gap between grammars and DNNs. It also enables flexible and diverse network structures to address Hinton's quest on improving structural sufficiency in DNNs~\cite{Hinton}.

\subsection{Method Overview}
We first summarize the best practices in existing building blocks, and then briefly overview our proposed AOG building block (Fig.~\ref{fig:AOG-block}) and how it unifies the existing ones. 

Existing building blocks usually do not fully implement the three principles (compositionality, reconfigurability and lateral connections).  
\begin{itemize} 
    \item InceptionNets or GoogLeNets~\cite{InceptionNet} embodies a split-transform-aggregate heuristic in a shallow feed-forward way for feature exploration, which is inspired by the network-in-network design~\cite{NetInNet} and the theoretical study on stage-wise design~\cite{BoundsOfDL}. However, the filter numbers and sizes are tailored for each individual transformation, and the modules are customized stage-by-stage. Interleaved group convolutions~\cite{IGC1} share the similar spirit, but use simpler scheme.
    \item ResNets~\cite{ResidualNet} provide a simple yet effective solution, inspired by the Highway network~\cite{HighwayNet}, that enables networks to enjoy going either deeper or wider without sacrificing the feasibility of optimization. From the perspective of representation learning, skip-connections within a ResNet~\cite{ResidualNet} contributes to effective feature reuse. They do not, however, realize the split component as done in GoogLeNets.
    \item ResNeXts~\cite{ResNeXt} add the spit component in ResNets and address the drawbacks of the Inception modules using group convolutions in the transformation.
    \item Deep Pyramid ResNets~\cite{DPRN} gradually increase feature channels between building blocks, instead of increasing feature channels sharply at each residual unit with down-sampling in vanilla ResNets.  
    \item DenseNets~\cite{DenseNet} explicitly differentiate between information that is added to the network and information that is preserved. Dense connections with feature maps being concatenated together are used, which are effective for feature exploration, but lack the capability of feature reuse as done in ResNets. 
    \item Dual Path Networks (DPN)~\cite{DPN} utilize ResNet blocks and DenseNet blocks in parallel to balance feature reuse and feature exploration. 
    \item Deep Layer Aggregation networks (DLA)~\cite{DLA} iteratively and hierarchically aggregate the feature hierarchy when stacking the building blocks such as the ResNet ones.  
\end{itemize}

Our AOG building block is hierarchical, compositional and reconfigurable with lateral connections by design. As Fig.~\ref{fig:AOG-block} shows, an AOG building block splits its input feature map into $N$ groups along feature channels, and treat it as a sentence of $N$ words. It then jointly realizes a phrase structure grammar (vertical composition)~\cite{Syntactic, Geman_CompositionSystems,DPM,Zhu_Grammar,Yuille_AndOr,DisAOT-CVPR} and a dependency grammar (horizontal connections in pink in Fig.~\ref{fig:AOG-block})~\cite{DependencyGrammar,Zhu_Grammar,RCN} in bottom-up parsing the ``sentence" for better feature exploration and reuse: 
\begin{itemize}
    \item Phrase structure grammar is a 1-D special case of the method presented in~\cite{DisAOT-CVPR, TLP-PAMI}. It can also be understood as a modified version of the well-known Cocke–-Younger–-Kasami (CYK) parsing algorithm in natural language processing according to a binary composition rule. \item Dependency grammar is integrated to capture lateral connections and improve the representational flexibility and power. 
\end{itemize}

In an AOG building block, each node applies some basic operation $\mathcal{T}(\cdot)$ (e.g., Conv-BN-ReLU) to its input, and there are three types of nodes: 
\begin{itemize} 
    \item A \textit{Terminal-node} takes as input a channel-wise slice of the input feature map (i.e., a $k$-gram). 
    \item An \textit{AND-node} implements composition, whose input is computed by concatenating features of its syntactic child nodes, and adding the lateral connection if present.
    \item An \textit{OR-node} represents alternative compositions, whose input is the element-wise sum of features of its syntactic child nodes and the lateral connection if present.
\end{itemize}

Our AOG building block unifies the best practices developed in popular building blocks in that,
\begin{itemize} 
    \item \textit{Terminal-nodes} implement the split-transform heuristic (or group convolutions) as done in GoogLeNets~\cite{InceptionNet} and ResNeXts~\cite{ResNeXt}, but at multiple levels (including overlapped group convolutions). They also implement the skip-connection at multiple levels. Unlike the cascade-based stacking scheme in ResNets, DenseNets and DPNs, Termninal-nodes can be computed in parallel to improve efficiency. Non-terminal nodes implement aggregation. 
    \item \textit{AND-nodes} implement DenseNet-like aggregation (i.e., concatenation)~\cite{DenseNet} for feature exploration. 
    \item \textit{OR-nodes} implement ResNet-like aggregation (i.e., summation)~\cite{ResidualNet} for feature reuse.
    \item \textit{The hierarchy} facilitates gradual increase of feature channels as in Deep Pyramid ResNets~\cite{DPRN}, and also leads to good balance between depth and width of networks. 
    \item \textit{The compositional structure} provides much more flexible information flows than DPN ~\cite{DPN} and the DLA~\cite{DLA}.
    \item \textit{The lateral connections} induce feature diversity and increase the effective depth of nodes along the path without introducing extra parameters. 
\end{itemize}

We stack AOG building blocks to form a deep AOG network, called \textbf{AOGNet}. Fig.~\ref{fig:model} illustrates a 3-stage AOGNet. Our AOGNet utilizes two nice properties of grammars: (i) The flexibility and simplicity of constructing different network structures based on a dictionary of primitives and a set of production rules in a principled way; and (ii) The highly expressive power and the parsimonious  compactness of their explicitly hierarchical and compositional structures.

\section{Related Work and Our Contributions}
Network architectures are the foundation for improving performance of DNNs. 
We focus on hand-crafted architectures in this section. Related work on neural architecture search is referred to the survey papers~\cite{NAS-survey1,NAS-survey2}.

\textbf{Hand-crafted network architectures.} 
After more than 20 years since the seminal work 5-layer LeNet5~\cite{LeCunCNN} was proposed, the recent resurgence in popularity of neural networks was triggered by the 8-layer AlexNet~\cite{AlexNet} with breakthrough performance on ImageNet~\cite{ImageNet} in 2012.
The AlexNet presented two new insights in the operator space: the Rectified Linear Unit (ReLu) and the Dropout. 
Since then, a lot of efforts were devoted to learn deeper AlexNet-like networks with the intuition that deeper is better. The VGG Net~\cite{VGG} proposed a 19-layer network with insights on using multiple successive layers of small filters (e.g., $3\times 3$) to obtain the receptive field by one layer with large filter and on adopting smaller stride in convolution to preserve information. 
A special case, $1\times 1$ convolution, was proposed in the network-in-network~\cite{NetInNet} for reducing or expanding feature dimensionality between consecutive layers, and have been widely used in many networks. The VGG Net also increased computational cost and memory footprint significantly.

The  22-layer GoogLeNet~\cite{GoogLeNet} introduced the first inception module and a bottleneck scheme implemented with $1\times 1$ convolution for reducing computational cost. The main obstacle of going deeper lies in the gradient vanishing issue in optimization, which is addressed with a new structural design, short-path or skip-connection, proposed in the Highway network~\cite{HighwayNet} and popularized by the ResNets~\cite{ResidualNet}, especially when combined with the batch normalization (BN)~\cite{BatchNorm}. More than 100 layers are popular design in the recent literature~\cite{ResidualNet,InceptionNet}, as well as even more than 1000 layers trained on large scale datasets such as ImageNet~\cite{StochasticResNet,PolyNet}. The Fractal Net~\cite{FractalNet} and deeply fused networks~\cite{DFN} provided an alternative way of implementing short path for training ultra-deep networks without residuals. Complementary to going deeper, width matters in ResNets and inception based networks too~\cite{WideResNet,ResNeXt,IGC1}. Going beyond the first-order  skip-connections in ResNets, DenseNets~\cite{DenseNet} proposed a densely connected network architecture with concatenation scheme for feature reuse and exploration, and DPNs~\cite{DPN} proposed to combine residuals and densely connections in an alternating way for more effective feature exploration and reuse. DLA networks~\cite{DLA} further develop iterative and hierarchical  aggregation schema with very good performance obtained.  
Both skip-connection and dense-connection adapt the sequential architecture to directed and acyclic graph (DAG) structured networks, which were explored earlier in the context of recurrent neural networks (RNN)~\cite{DAGRNN,DAGRNN1} and ConvNets~\cite{DAGCNN}.  

Most work focused on boosting spatial encoding and utilizing spatial dimensionality reduction. The squeeze-and-excitation module~\cite{SENet} is a recently proposed simple yet effective method focusing on channel-wise encoding. The Hourglass network~\cite{Hourglass} proposed a hourglass module consisting of both subsampling and upsampling to enjoy repeated bottom-up/top-down feature exploration.

Our AOGNet is created by intuitively simple yet principled grammars. It shares some spirit with the inception module~\cite{InceptionNet}, the deeply fused nets~\cite{DFN} and the DLA~\cite{DLA}.

\textbf{Grammars.} A general framework of image grammar was proposed in~\cite{Zhu_Grammar}. Object detection grammar was the dominant approaches for object detection~\cite{Pff_Grammar,Yuille_AndOr,DisAOT-CVPR,StochasticGrammar,AOG_Action,AOG_Shape}, and has recently been integrated with DNNs~\cite{YingWu_CompositionalPose,YingWu_CompositionalModel,AOG_TaskPlanning}. Probabilistic program induction~\cite{ProbabilisticPrograming,lake15science,BuildMachineLikePeople} has been used successfully in many settings, but has not shown good performance in difficult visual understanding tasks such as large-scale image classification and object detection. More recently, recursive cortical networks~\cite{RCN} have been proposed with better data efficiency in learning which adopts the AND-OR grammar framework~\cite{Zhu_Grammar}, showing great potential of grammars in developing general AI systems. 

\textbf{Our contributions.} This paper makes two main contributions in the field of deep representation learning: 
\begin{itemize} 
    \item It proposes compositional grammatical architectures for deep learning and presents deep AND-OR Grammar networks (AOGNets). AOGNets facilitate both feature exploration and feature reuse in a hierarchical and compositional way. AOGNets unify the best practices developed by state-of-the-art DNNs such as GoogLeNets, ResNets, ResNeXts, DenseNets, DPN, and DLA. To our best knowledge, it is the first work of designing grammar-guided network generators. 
    \item It obtains better performance than many state-of-the-art networks in the CIFAR and ImageNet-1K classification benchmark and MS-COCO object detection and segmentation benchmark. It also obtains better model interpretability  and shows greater potential for adversarial defense.
\end{itemize}

\section{AOGNets} \label{sec:AOGNets}
In this section, we first present details of constructing the structure of our AOGNets. Then, we define node operation functions for nodes in an AOGNet. We also propose a method of simplifying the full structure of an AOG building block which prunes syntactically symmetric nodes.  

\subsection{The Structure of an AOGNet} \label{sec:structure}
An AOGNet (Fig.~\ref{fig:model}) consists of a predefined number of stages each of which comprises one or more than one AOG building blocks.
As Fig.~\ref{fig:AOG-block} illustrates, an AOG building block maps an input feature map $F_{in}$ with the dimensions  $D_{in}\times H_{in}\times W_{in}$ (representing the number of channels, height and width respectively) to an output feature map $F_{out}$ with the dimensions $D_{out}\times H_{out}\times W_{out}$. \emph{We split the input feature map into $N$ groups along feature channels, and then treat it as a ``sentence of $N$ words".} Each ``word" represents a slice of the input feature map with $\frac{D_{in}}{N}\times H_{in} \times W_{in}$. 
Our AOG building block is constructed by a simple algorithm (Algorthm~\ref{alg:AOG}) which integrates two grammars. 

\textbf{The phrase structure grammar}~\cite{Syntactic,Geman_CompositionSystems,DPM,Zhu_Grammar,Yuille_AndOr,DisAOT-CVPR}. Let $S_{i,j}$ be a non-terminal symbol representing the sub-sentence starting at the $i$-th word ($i\in [0, N-1]$) and ending at the $j$-th word ($j\in[0, N-1], j\geq i$) with the length $k=j-i+1$. We consider the following three rules in parsing a sentence: 
\begin{align}
    S_{i,j}  \rightarrow & \quad t_{i,j}, \\
    S_{i,j}(m)  \rightarrow & \quad [L_{i, i+m}\cdot R_{i+m+1, j}], \quad 0\leq m <k, \\
    S_{i,j} \rightarrow & \quad S_{i,j}(0) | S_{i,j}(1) | \cdots | S_{i,j}(j-i).
\end{align}
where we have,
\begin{itemize} 
    \item The first rule is a termination rule which grounds the non-terminal symbol $S_{i,j}$ directly to the corresponding sub-sentence $t_{i,j}$, i.e., a $k$-gram terminal symbol, which is represented by a \textbf{ Terminal-node}. 
    \item The second rule is a binary decomposition rule, denoted by $[ L \cdot R]$, which decomposes a non-terminal symbol $S_{i,j}$ into two child non-terminal symbols representing a left sub-sentence and a right sub-sentence, $L_{i,i+m}$ and $R_{i+m+1,j}$ respectively. It is represented by an \textbf{AND-node}, and entails \textbf{the concatenation scheme} in forward computation to match feature channels. 
    \item The third rule represents alternative ways of decomposing a non-terminal symbol $S_{i,j}$, denoted by $A | B | C$, which is represented by an \textbf{OR-node}, and can utilize \textbf{summation scheme} in forward computation to ``integrate out" the decomposition structures. 
\end{itemize}

\begin{algorithm} 
\SetAlgoLined
\KwIn{The total length (or primitive size) $N$. 
}
\KwOut{The AND-OR Graph $\mathcal{G}=<V, E>$}
Initialization: Create an OR-node $O_{0,N-1}$ for the entire sentence, $V=\{O_{0,N-1}\}, E=\emptyset$, BFS queue $Q=\{O_{0,N-1}\}$\;
\While{$Q$ is not empty}{
Pop a node $v_{i,j}$ from the $Q$ and let $k=j-i+1$\; 
\uIf{$v_{i,j}$ is an OR-node}{
      	i) Add a terminal-node $t_{i,j}$, and update $V=V\cup\{t_{i,j}\},\, E=E\cup\{<v_{i,j}, t_{i,j}>\}$\;
      	ii) Create AND-nodes $A_{i,j}(m)$ for all valid splits $0\leq m<k$\;
      	$E=E\cup\{<v_{i,j}, A_{i,j}(m)>\}$\;
      	\If{$A_{i,j}(m)\notin V$} {
      		$V=V\cup\{ A_{i,j}(m) \}$\; 
      		Push $A_{i,j}(m)$ to the back of $Q$\;
      	}
}
\ElseIf{$v_{i,j}$ is an AND-node with split index $m$} {
      	Create two OR-nodes $O_{i,i+m}$ and $O_{i+m+1, j}$ for the two sub-sentence respectively\;
      	$E=E\cup\{<v_{i,j}(m), O_{i,i+m}>, <v_{i,j}(m), O_{i+m+1,j}>\}$\;
      	\If{$O_{i,i+m}\notin V$} {
      		$V=V\cup\{ O_{i,i+m} \}$\;
      		Push $O_{i,i+m}$ to the back of $Q$\;
      	}
      	\If{$O_{i+m+1,j}\notin V$} {
      		$V=V\cup\{ O_{i+m+1, j} \}$\;
      		Push $O_{i+m+1,j}$ to the back of $Q$\;
      	}
}  
}
Add lateral connections (see text for detail).  
\caption{Constructing an AOG building block}\label{alg:AOG} 
\end{algorithm}

\textbf{The dependency grammar}~\cite{DependencyGrammar,RCN,Zhu_Grammar}. We introduce dependency grammar to model lateral connections between non-terminal nodes of the same type (AND-node or OR-node) with the same length $k$. As illustrated by the arrows in pink in Fig.~\ref{fig:AOG-block}, \textit{we add lateral connections in a straightforward way}: (i) For the set of OR-nodes with $k\in [1, N-1]$, we first sort them based on the starting index $i$; and (ii) For the set of AND-nodes with $k\in [2, N]$, we first sort them based on the lexical orders of the pairs of starting indexes of the two child nodes. Then, we add sequential lateral connections for nodes in the sorted set either from left to right, or vice versa. We use opposite lateral connection directions for AND-nodes and OR-nodes iteratively to have globally consistent lateral flow from bottom to top in an AOG  building block.

\subsection{Node Operations in an AOGNet}\vspace{-1mm}
In an AOG building bock, all nodes use the same type of transformation function $\mathcal{T}(\cdot)$ (see Fig.~\ref{fig:AOG-block}). For a node $v$, denote by $f_{in}(v)$ its input feature map,  and then its output feature map is computed by $f_{out}(v)=\mathcal{T}(f_{in}(v))$. For a Terminal-node $t$, it is straightforward to apply the transformation using $f_{in}(t)=F_{in}(t)$ where $F_{in}(t)$ is the $k$-gram slice from the input feature map of the AOG building block. For AND-nodes and OR-nodes, we have, 
\begin{itemize}
    \item For an AND-node $A$ with two child nodes $L$ and $R$, its input $f_{in}(A)$ is first computed by the concatenation of the outputs of the two child nodes, $f_{in}(A)=[f_{out}(L)\cdot \lambda_L, f_{out}(R)\cdot \lambda_R]$. If it has a lateral node whose output is denoted by $f_{out}(A_{lateral})$, we add it and get  $f_{in}(A)=[f_{out}(L)\cdot \lambda_L, f_{out}(R)\cdot \lambda_R]+f_{out}(A_{lateral})\cdot \lambda_{lateral}$. 
    \item For an OR-node $O$, its input  is the summation of the outputs of its child nodes (including the lateral  node if present), $f_{in}(O)=\sum_{u\in ch(O)} f_{out}(u)\cdot \lambda_u$, where $ch(\cdot)$ represents the set of child nodes.  
\end{itemize}
Where $\lambda_L, \lambda_R$, $\lambda_{lateral}$ and $\lambda_u$'s are weights (see details in Section~\ref{sec:detail}). Node inputs are computed following the syntactical structure of AOG building block to ensure that feature dimensions and spatial sizes match in the concatenation and summation.   
In learning and inference, we follow the depth-first search (DFS) order to compute nodes in an AOG building block, which ensures that all the child nodes have been computed when we compute a node $v$.

\begin{figure} [t]
    	\centering
    	\includegraphics[width = 0.5\textwidth]{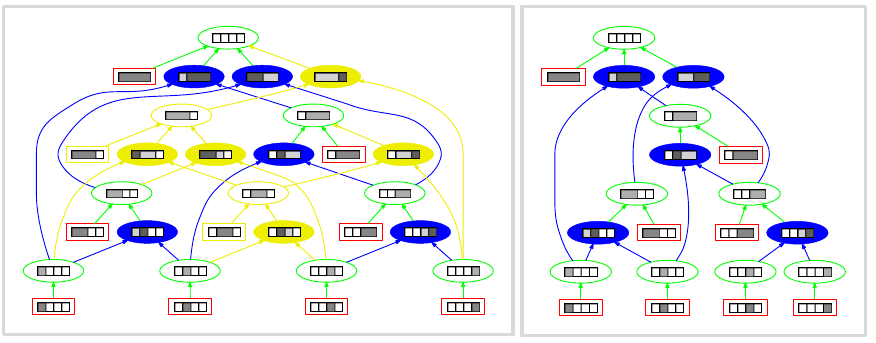}
    	\caption{\small{Illustration of simplifying the AOG building blocks by pruning syntactically symmetric child nodes of OR-nodes. \textit{Left:} An AOG building block with full structure consisting of  $10$ Terminal-nodes, $10$ AND-nodes and $10$ OR-nodes. Nodes and edges to be pruned are plotted in yellow. \textit{Right:} The simplified AOG building block consisting of  $8$ Terminal-nodes, $5$ AND-nodes and $8$ OR-nodes. (Best viewed in color)}
    	}
    	\label{fig:pruning} 
\end{figure}

\subsection{Simplifying AOG Building Blocks} \vspace{-1mm}
The phrase structure grammar is syntactically redundant since it unfolds all possible configurations w.r.t. the binary composition rule. In representation learning, we also want to increase the feature dimensions of different stages in a network for better representational power, but try not to increase the total number of parameters significantly. \textbf{To balance the structural complexity and the feature dimensions} of our AOG building block, we propose to simplify the structure of an AOG building block by pruning some syntactically redundant nodes. As illustrated in Fig.~\ref{fig:pruning}, the pruning algorithm is simple: Given a full structure AOG building block, we start with an empty simplified block. We first add the root OR-node into the simplified block. Then, we follow the BFS order of nodes in the full structure block. For each encountered OR-node we only keep the child nodes which do not have left-right syntactically symmetric counterparts in the current set of child nodes in the simplified block. For encountered AND-nodes and Terminal-nodes, we add them to the simplified block. The pruning algorithm can be integrated into Algorithm~\ref{alg:AOG}. For example, consider the four child nodes of the root OR-node in the left of Fig.~\ref{fig:pruning}, the fourth child node is removed since it is symmetric to the second one.

\section{Experiments}\label{sec:Exp}
Our AOGNet is tested in the CIFAR-10 and CIFAR-100~\cite{CIFAR}, and ImageNet-1K~\cite{ImageNet} classification benchmark  and the MS-COCO object detection and segmentation benchmark~\cite{COCO}.

\subsection{Implementation Settings and Details}\label{sec:detail}
We use simplified AOG building blocks. For the node operation $\mathcal{T}()$, we use the bottleneck variant of Conv-BN-ReLU proposed in ResNets~\cite{ResidualNet}, which adds one $1\times 1$ convolution before and after the operation to first reduce feature dimension and then expand it back. More specifically, we have $\mathcal{T}(x) = ReLU(x + T(x))$ for an input feature map $x$ where $T()$ represents a sequence of primitive operations, Conv1x1-BN-ReLU, Conv3x3-BN-ReLU and Conv1x1-BN. If Dropout~\cite{AlexNet} is used with drop rate $p\in (0, 1)$, we add it after the last BN, i.e., $\mathcal{T}(x) = ReLU(x + Dropout(T(x),p))$

\textit{Handling double-counting due to  the compositional DAG structure and lateral connections.} First, in our AOG building block, some nodes will have multiple paths to reach the root OR-node due to the compositional DAG structure. Since we use the skip connection in the node operation $\mathcal{T}()$, the feature maps of those nodes with multiple paths will be double-counted at the root OR-node. Second, if a node $v$ and its lateral node $v_{lateral}$ share a parent node, we also need to handle double-counting in the skip connection. Denote by $n(v)$ the number of paths between $v$ and the root OR-node, which can be counted during the building block construction (Algorithm~\ref{alg:AOG}). Consider an AND-node $A$ with two syntactic child node $L$ and $R$ and the lateral node $A_{lateral}$, we compute two different inputs, one for the skip connection, $f^{skip}_{in}(A) = [f_{out}(L)\cdot \frac{n(A)}{n(L)},f_{out}(R)\cdot \frac{n(A)}{n(R)}]$ if $A$ and $A_{lateral}$ share a parent node and $f^{skip}_{in}(A) = [f_{out}(L)\cdot \frac{n(A)}{n(L)},f_{out}(R)\cdot \frac{n(A)}{n(R)}]+f_{out}(A_{lateral})\cdot \frac{n(A)}{n(A_{lateral})}$ otherwise, and the other for $T()$, $f^T_{in}(A)=[f_{out}(L), f_{out}(R)]+f_{out}(A_{lateral})$. The transformation for node $A$ is then implemented by $\mathcal{T}(A)=ReLU(f^{skip}(A)+T(f^T_{in}(A)))$. Similarly, we can set $\lambda_u$'s in the OR-node operation. We note that we can also treat $\lambda$'s as unknown parameters to be learned end-to-end.

\subsection{Image Classification in ImageNet-1K} 
The ILSVRC 2012 classification dataset~\cite{ImageNet} consists of about $1.2$ million images for training, and $50,000$ for validation, from $1,000$ classes. We adopt the same data augmentation scheme (random crop and horizontal flip) for training images as done in~\cite{ResidualNet, DenseNet}, and apply a single-crop with size $224\times224$ at test time. Following the common protocol,  we evaluate the top-1 and top-5 classification error rates on the validation set.

\textbf{Model specifications.} We test three AOGNets with different model complexities. In comparison, we use the model size as the name tag for AOGNets (e.g., AOGNet-12M means the AOGNet has 12 million parameters or so). The stem (see Fig.~\ref{fig:model}) uses three Conv3x3-BN layers  (with stride $2$ for the first layer), followed by a $2\times2$ max pooling layer with stride $2$. 
All the three AOGNets use four stages. Within a stage, we use the same AOG building block, while different stages may use different blocks. A stage is then specified by $N_n$ where $N$ is primitive size (Algorithm~\ref{alg:AOG}) and $n$ the number of blocks. The filter channels are defined by a 5-tuple for specifying the input and output dimensions for the 4 stages. 
The detailed specifications of the three AOGNets are: AOGNet-12M uses stages of $(2_2, 4_1, 4_3, 2_1)$ with filter channels $(32, 128, 256, 512, 936)$, AOGNet-40M uses stages of $(2_2, 4_1, 4_4, 2_1)$ with filter channels $(60, 240, 448, 968, 1440)$, and AOGNet-60M uses stages of $(2_2, 4_2, 4_5, 2_1)$ withe filter channels $(64, 256, 512, 1160, 1400)$. 

\begin{table}
    \centering
    \small{
    \resizebox{0.48\textwidth}{!}{
    \begin{tabular}{c|l|l|l|l}
    \hline 
    Method & \#Params & FLOPS & top-1 & top-5 \\ \hline
    ResNet-101~\cite{ResidualNet} & 44.5M & 8G & 23.6 & 7.1 \\ 
    ResNet-152~\cite{ResidualNet} & 60.2M & 11G & 23.0 & 6.7 \\ \hline
    ResNeXt-50~\cite{ResNeXt} & 25.03M & 4.2G & 22.2 & 5.6 \\ 
    ResNeXt-101 (32$\times$ 4d)~\cite{ResNeXt} & 44M & 8.0G & 21.2 & 5.6 \\
    ResNeXt-101 (64$\times$ 4d)~\cite{ResNeXt} & 83.9M & 16.0G & 20.4 & 5.3 \\ 
    ResNeXt-101 + BAM~\cite{park2018bam} & 44.6M & 8.05G & 20.67 & - \\ 
    ResNeXt-101 + CBAM~\cite{woo2018cbam} & 49.2M & 8.0G & 20.60 & - \\ 
    ResNeXt-50+SE~\cite{SENet} & 27.7M & 4.3G & 21.1 & 5.49 \\
    ResNeXt-101+SE~\cite{SENet} & 48.9M & 8.46G & 20.58 & 5.01 \\\hline
    DensetNet-161~\cite{DenseNet}  & 27.9M & 7.7G & 22.2 & - \\ 
    DensetNet-169~\cite{DenseNet}  & $\sim$ 13.5M & $\sim$ 4G & 23.8 & 6.85 \\
    DensetNet-264~\cite{DenseNet}  & $\sim$ 33.4M & - & 22.2 & 6.1 \\ 
    DensetNet-cosine-264~\cite{DenseNet-efficient}  & $\sim$ 73M & $\sim$ 26G & 20.4 & - \\ \hline
    DPN-68~\cite{DPN} & 12.8M & 2.5G & 23.57 & 6.93 \\ 
    DPN-92~\cite{DPN} & 38.0M & 6.5G & 20.73 & 5.37 \\
    DPN-98~\cite{DPN} & 61.6M & 11.7G & 20.15 & 5.15 \\ \hline
    {AOGNet-12M} & {11.9M} & {2.36G} & {22.28} & {6.14}  \\ 
    {AOGNet-40M} & 40.3M & 8.86G & {19.82} & {4.88} \\ 
    {AOGNet-60M} & 60.7M & 14.36G & \textbf{19.34} & \textbf{4.78}  \\ \hline
    \end{tabular} }}
    \\ [1ex]
    \caption{The top-1 and top-5 error rates (\%) on the ImageNet-1K validation set using single model and single-crop testing. 
    }\label{table:imagenet-results} 
\end{table}

\begin{figure} 
    	\centering
    	\includegraphics[width = 0.5\textwidth]{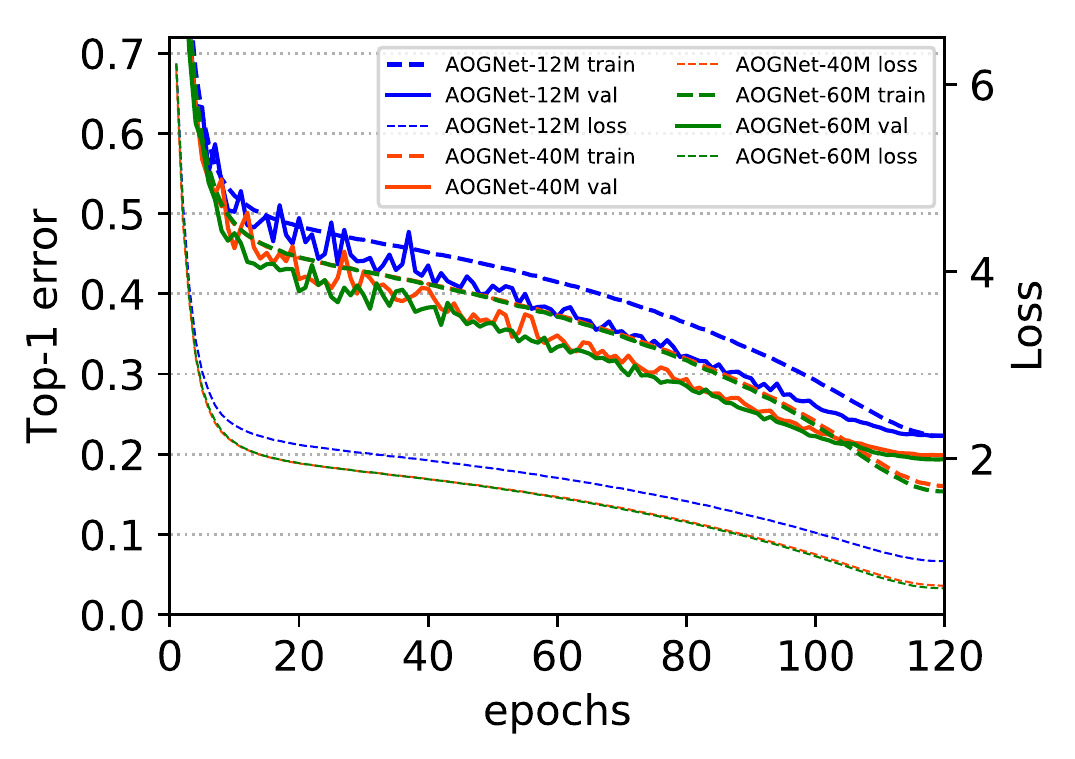}
    	\caption{\small{Plots of  top-1 error rates and  training losses of the three AOGNets in ImageNet. (Best viewed in color and magnification)}
    	}
    	\label{fig:plots} 
\end{figure}

\begin{figure} 
    	\centering
    	\includegraphics[width = 0.5\textwidth]{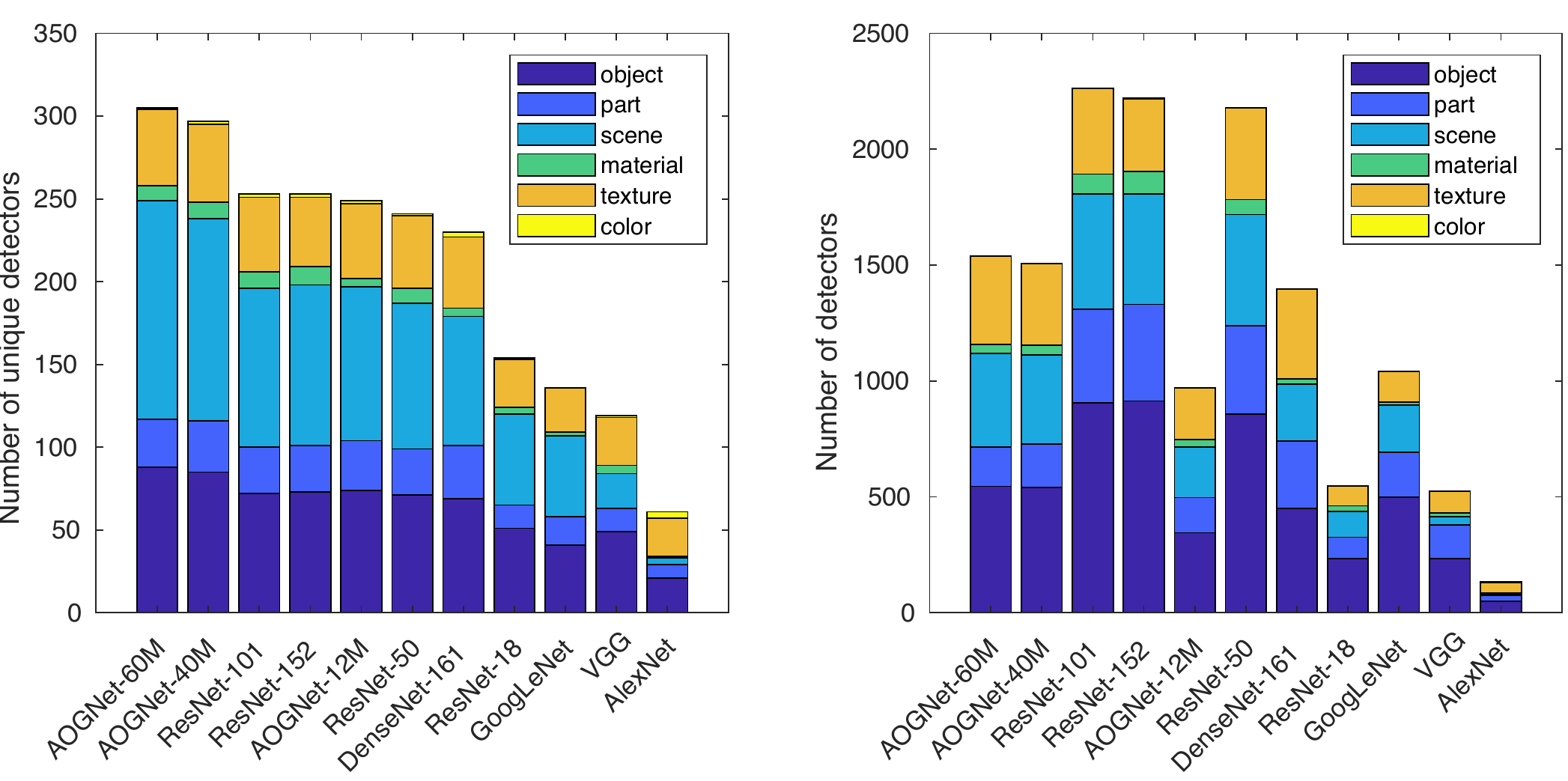}
    	\caption{\small{Comparisons of model interpretability using the \href{http://netdissect.csail.mit.edu/}{network dissection} method~\cite{netdissect2017} on ImageNet pretrained networks.}
    	}
    	\label{fig:dissection} 
\end{figure}

\textbf{Training settings.} We adopt random parameter initialization for filter weights. For Batch Normalization (BN) layers, we use $0$ to initialize all offset parameters. We use $1$ to initialize all scale parameters except for the last BN layer in each $\mathcal{T}()$ where we initialize the scale parameter by $0$ as done in~\cite{TrainImageNet1Hour}. We use Dropout~\cite{AlexNet} with drop rate $0.1$ in the last two stages. 
We use 8 GPUs (NVIDIA V100) in training. The batch size is $128$ per GPU ($1024$ in total). The initial learning rate is  $0.4$, and the cosine learning rate scheduler~\cite{cosine_lr} is used with weight decay $1\times10^{-4}$ and momentum $0.9$. We train AOGNet with SGD for $120$ epochs which include $5$ epochs for linear warm-up following~\cite{TrainImageNet1Hour}.

\begin{table} 
    \centering
    \small{
    \resizebox{0.45\textwidth}{!}{
    \begin{tabular}{lcccc}
    \hline 
    Method & \#Params & $\epsilon=0.1$ & $\epsilon=0.3$ & clean \\ \hline
    ResNet-101 & 44.5M & 12.3 & 0.40 & 77.37 \\ 
    ResNet-152 & 60.2M & 16.3 & 0.85 & 78.31 \\ 
    DenseNet-161 & 28.7M & 13.0 & 2.1 & 77.65 \\ 
    AOGNet-12M & 12.0M & 18.1 & 1.4 & 77.72 \\ 
    AOGNet-40M & 40.3M & 28.3 & 2.2 & 80.18 \\ 
    AOGNet-60M & 60.1M & 30.2 & 2.6 & 80.66 \\ \hline
    \end{tabular} }}
    \\ [1ex]
    \caption{Top-1 accuracy comparisons under white-box adversarial attack using 1-step FGSM~\cite{FGSM} with the \href{https://github.com/bethgelab/foolbox}{Foolbox} toolkit~\cite{Foolbox}.  
    }\label{table:attack-results} 
\end{table}

\textbf{Results and Analyses: AOGNets obtain the best accuracy and model interpretability.}  Table~\ref{table:imagenet-results} shows the results, and Fig.~\ref{fig:plots} shows plots for the top-1 error rates and training losses. 
Our AOGNets are the best among the models with comparable model sizes in comparison in terms of top-1 and top-5 accuracy. 
Our small AOGNet-12M outperforms ResNets~\cite{ResidualNet} ($44.5M$ and $60.2M$) by $1.32\%$ and $0.72\%$ respectively. \textit{We note that our AOGNets use the same bottleneck operation function as ResNets, so the improvement must be contributed by the AOG building block structure.} Our AOGNet-40M obtains better performance than all other methods in comparison, including ResNeXt-101~\cite{ResNeXt}+SE~\cite{SENet} ($48.9M$) which represents the most powerful and widely used combination in practice. AOGNet-40M also obtains better performance than the runner-up, DPN-98~\cite{DPN} ($61.6M$), which indicates that the hierarchical and compositional integration of the DenseNet- and ResNet-aggregation in our AOG building block is more effective than the cascade-based integration in the DPN~\cite{DPN}. Our AOGNet-60M achieves the best results. The FLOPs of our AOGNet-60M are slightly higher than DPN-98 partially because DPN uses ResNeXt operation (i.e., group conv.). In our on-going experiments, we are testing AOGNets with ResNeXt node operation. 

\textit{Model Interpretability} has been recognized as a critical concern in developing deep learning based AI systems~\cite{XAI}. We use the network dissection metric~\cite{netdissect2017} which compares the number of unique ``detectors" (i.e., filter kernels) in the last convolution layer. Our AOGNet obtains the best score in comparison (Fig.~\ref{fig:dissection}), which indicates the AOG building block has great potential to induce model interpretabilty by design, while achieving the best accuracy performance.  

\textit{Adversarial robustness} is another crucial issue faced by many DNNs~\cite{AdversarialExampl}. We conduct a simple experiment to compare the out-of-the-box adversarial robustness of different DNNs. Table~\ref{table:attack-results} shows the results. Under the vanilla settings, our AOGNets show better potential in adversarial defense, especially when the perturbation energy is controlled relatively low (i.e. $\epsilon=0.1$). We will investigate this with different attacks and adversarial training in future work. 

\textit{Mobile settings.} We train an AOGNet-4M under the typical mobile settings~\cite{howard2017mobilenets}. Table~\ref{table:imagenet-results-small} shows the comparison results. We obtain performance on par to the popular networks specifically designed for mobile platforms such as the MobileNets~\cite{howard2017mobilenets,sandler2018mobilenetv2} and ShuffleNets~\cite{zhang2017shufflenet}. Our AOGNet also outperforms the auto-searched network, NASNet~\cite{zoph2017learning} (which used around 800 GPUs in search). \textit{We note that we use the same AOGNet structure, thus showing promising device-agnostic capability of our AOGNets.} This is potentially  important and useful for deploying DNNs to different platforms in practice since no extra efforts of hand-crafting or searching neural architectures are entailed. This will be also potentially useful for distilling a small model from a large model if they share the exactly same structure.

\begin{table}
    \centering
    \small{
    \resizebox{0.48\textwidth}{!}{
    \begin{tabular}{c|c|c|c|c}
    \hline 
    Method & \#Params & FLOPS & top-1 & top-5 \\ \hline
    MobileNetV1~\cite{howard2017mobilenets} & 4.2M & 575M & 29.4 & 10.5 \\ 
    SqueezeNext~\cite{gholami2018squeezenext} & 4.4M & - & 30.92 & 10.6  \\
    ShuffleNet (1.5)~\cite{zhang2017shufflenet} & 3.4M & 292M & 28.5 & - \\ 
    ShuffleNet (x2)~\cite{zhang2017shufflenet} & 5.4M & 524M & 26.3 & - \\ 
    CondenseNet (G=C=4)~\cite{huang2017condensenet} & 4.8M & 529M & 26.2 & 8.3 \\ 
    MobileNetV2~\cite{sandler2018mobilenetv2} & 3.4M & 300M & 28.0 & 9.0 \\ 
    MobileNetV2 (1.4)~\cite{sandler2018mobilenetv2} & 6.9M & 585M & 25.3 & 7.5 \\ 
    NASNet-C (N=3)~\cite{zoph2017learning} & 4.9M & 558M & 27.5 & 9.0 \\ \hline
    {AOGNet-4M} & 4.2M & 557M & {26.2} & {8.24}  \\ \hline
    \end{tabular} }}
    \\ [1ex]
    \caption{The top-1 and top-5 error rates (\%) on the ImageNet-1K validation set under mobile settings. 
    }\label{table:imagenet-results-small} 
\end{table}

\begin{table} [t]
    \centering
    \small{
    \resizebox{0.5\textwidth}{!}{
    \begin{tabular}{lcc|ccc|ccc}
    \hline 
    Method & \#Params & $t$ (s/img) & AP$^{bb}$ & AP$^{bb}_{50}$ & AP$^{bb}_{75}$ & AP$^{m}$ & AP$^{m}_{50}$ & AP$^{m}_{75}$ \\ \hline
    ResNet-50-C4 & 35.9M & 0.130 & 35.6 & 56.1 & 38.3 & 31.5 & 52.7 & 33.4 \\ 
    ResNet-101-C4 & 54.9M & 0.180 & 39.2 & 59.3 & 42.2 & 33.8 & 55.6 & 36.0  \\ 
    AOGNet-12M-C4 & 14.6M & 0.092& 36.8 & 56.3 & 39.8 & 32.0 & 52.9 & 33.7\\
    AOGNet-40M-C4 & 48.1M & 0.184 & \textbf{41.4} & \textbf{61.4} & \textbf{45.2} & \textbf{35.5} & \textbf{57.8} & \textbf{37.7} \\ \hline
    ResNet-50-FPN & 44.3M & 0.125 & 37.8 & 59.2 & 41.1 & 34.2 & 56.0 & 36.3 \\ 
    ResNet-101-FPN & 63.3M & 0.145 & 40.1 & 61.7 & 44.0 & 36.1 & 58.1 & 38.3 \\
    ResNeXt-101-FPN & 107.4M & 0.202 & 42.2 & 63.9 & 46.1 & 37.8 & 60.5 & 40.2\\
    AOGNet-12M-FPN & 31.2M & 0.122 & 38.0 & 59.8 & 41.3 & 34.6 & 56.6 & 36.4 \\ 
    AOGNet-40M-FPN & 59.4M & 0.147 & 41.8 & 63.9 & 45.7 & 37.6 & 60.3 & 40.1 \\
    AOGNet-60M-FPN & 78.9M & 0.171 & \textbf{42.5} & \textbf{64.4} & \textbf{46.7} & \textbf{37.9} & \textbf{60.9} & \textbf{40.3} \\
    \hline
    \end{tabular} }}
    \\ [1ex]
    \caption{\small{Mask-RCNN results on \textit{coco\_val2017} using the 1x training schedule. Results of ResNets and ResNeXts are reported by the \href{https://github.com/facebookresearch/maskrcnn-benchmark}{maskrcnn-benchmark}. }
    }\label{table:coco-results} 
\end{table}

\subsection{Object Detection and Segmentation in COCO}
MS-COCO is one the widely used benchmarks for object detection and segmentation~\cite{COCO}. It consists of 80 object categories. We train AOGNet in the COCO \texttt{train2017} set and evaluate in the COCO \texttt{val2017} set. We report the standard COCO metrics of Average Precision (AP), AP$_{50}$, and AP$_{75}$, for bounding box detection (AP$^{bb}$) and instance segmentation, i.e. mask prediction (AP$^m$).
We experiment on the Mask-RCNN system~\cite{MaskRCNN} using the state-of-the-art implementation, \texttt{maskrcnn-benchmark}~\cite{maskrcnn-code}. 
We use AOGNets pretrained on ImageNet-1K as the backbones. In fine-tuning for object detection and segmentation, we freeze all the BN parameters as done for the ResNet~\cite{ResidualNet} and ResNeXt~\cite{ResNeXt} backbones. We keep all remaining aspects unchanged. We test both the \texttt{C4} and \texttt{FPN} settings. 

\textbf{Results.} Table~\ref{table:coco-results} shows the comparison results. Our AOGNets  obtain better results than the ResNet~\cite{ResidualNet} and ResNeXt~\cite{ResNeXt} backbones with smaller model sizes and similar or slightly better inference time. The results show the effectiveness of our AOGNets learning better features in object detection and segmentation tasks.

\subsection{Experiments on CIFAR}
CIFAR-10 and CIFAR-100 datasets~\cite{CIFAR}, denoted by C10 and C100 respectively,  consist of $32\times32$ color images drawn from 10 and 100 classes. The training and test sets contains $50,000$ and $10,000$ images respectively. We adopt widely used standard data augmentation scheme, random cropping and mirroring, in preparing the training data. 

We train AOGNets with stochastic gradient descent (SGD) for $300$ epochs with random parameter initialization. The front-end (see Fig.~\ref{fig:model}) uses a single convolution layer. The initial learning rate is set to $0.1$, and is divided by $10$ at $150$ and $225$ epoch respectively. For CIFAR-10, we chose batch size $64$ with weight decay $1\times10^{-4}$, while batch size $128$ with weight decay $5\times10^{-4}$ is adopted for CIFAR-100. The momentum is set to $0.9$.

\textbf{Results and Analyses.} We summarize the results in Table~\ref{table:cifar-results}. With smaller model sizes and much reduced computing complexity (FLOPs), our AOGNets obtain better performance than ResNets~\cite{ResidualNet} and some of the variants, ResNeXts~\cite{ResNeXt} and DenseNets~\cite{DenseNet} consistently on both datasets. Our small AOGNet ($0.78M$) already outperforms the ResNet~\cite{ResidualNet} ($10.2M$) and the WideResNet~\cite{WideResNet} ($11.0M$). Since the same node operation is used, the improvement must come from the AOG building block structure. Compared with the DenseNets, our AOGNets improve more on C100, and use less than half FLOPs for comparable model sizes. The reason for the reduced FLOPs is that DenseNets apply down-sampling after each Dense block, while our AOGNets sub-sample at Terminal-nodes.  

\subsection{Ablation Study}
We conduct an ablation study which investigates the effects of \emph{(i) RS:} Removing Symmetric child nodes of OR-nodes in the pruned AOG building blocks, and of \emph{(ii) LC:} adding Lateral Connections. As Table~\ref{table:ablation-cifar} shows, the two components, RS and LC, improve performance. The results are consistent with our design intuition and principles. The RS component facilitates higher feature dimensions due to the reduced structural complexity, and the LC component increases the effective depth of nodes on the lateral flows.  

\begin{table} [t]
        \centering 
        \small{
        \resizebox{0.48\textwidth}{!}{
        \begin{tabular}{c|ccc|cc}
        \hline 
        Method & Depth & \#Params & FLOPs & C10 & C100\\ \hline
        ResNet~\cite{ResidualNet} & 110 & 1.7M & 0.251G & 6.61 & - \\ \hline
        ResNet (reported by~\cite{StochasticResNet}) & 110 & 1.7M & 0.251G & 6.41 & 27.22 \\ \hline
        \multirow{2}{*}{ResNet (pre-activation)~\cite{ResNetPreAct}} & 164 & 1.7M & 0.251G & 5.46 & 24.33 \\ 
        & 1001 & 10.2M & - & 4.62 & 22.71 \\ \hline
        Wide ResNet~\cite{WideResNet} & 16 & 11.0M & -& 4.81 & 22.07 \\ \hline
        DenseNet-BC~\cite{DenseNet} ($k = 12$) & 100 & 0.8M & 0.292G & 4.51 & 22.27 \\ \hline
        \textbf{AOGNet-1M} & - & \textbf{0.78M} & \textbf{0.123G} & \textbf{4.37} & \textbf{20.95} \\ \Xhline{2\arrayrulewidth}
        
        DenseNet-BC~\cite{DenseNet} ($k = 24$) & 250 & 15.3M & 5.46G &  3.62 & 17.60\\ \hline
		\textbf{AOGNet-16M} & - & 15.8M & \textbf{2.4G} & \textbf{3.42} & \textbf{16.93} \\ \Xhline{2\arrayrulewidth}

        Wide ResNet~\cite{WideResNet} & 28 & 36.5M & 5.24G & 4.17 & 20.50 \\ \hline
        FractalNet~\cite{FractalNet} & 21 & 38.6M &- & 5.22 & 23.30 \\
        with Dropout/DropPath & 21 & 38.6M &- & 4.60 & 23.73 \\ \hline
        ResNeXt-29, $8\times64$d~\cite{ResNeXt} & 29 & 34.4M & 3.01G & 3.65 & 17.77 \\ 
        ResNeXt-29, $16\times64$d~\cite{ResNeXt} & 29 & 68.1M & 5.59G & 3.58 & 17.31 \\ \hline
		
		DenseNet-BC~\cite{DenseNet} ($k = 40$) & 190 & 25.6M & 9.35G & 3.46 & 17.18 \\ \hline
		\textbf{AOGNet-25M} & - & \textbf{24.8M} & 3.7G & \textbf{3.27} & \textbf{16.63} \\ \hline 
	
        \end{tabular} } }
        \\ [1ex]
        \caption{Error rates (\%) on the two CIFAR datasets~\cite{CIFAR}. \#Params uses the unit of Million. $k$ in DenseNet refers to the growth rate. 
        }\label{table:cifar-results} 
    \end{table}

 \begin{table}
    \centering
    \small{
    \resizebox{0.45\textwidth}{!}{
    \begin{tabular}{c|c|c|c|c}
    \hline 
    Method & \#Params & FLOPS & CIFAR10 & CIFAR100 \\ \hline
    AOGNet & 4.24M & 0.65G & 3.75 & 19.20 \\ 
    AOGNet+LC & 4.24M & 0.65G & 3.70 & 19.09 \\ 
    AOGNet+RS & 4.23M & 0.70G & 3.57 & 18.64 \\
    AOGNet+RS+LC & 4.23M & 0.70G & \textbf{3.52} & \textbf{17.99} \\ \hline
    \end{tabular} }}
    \\ [1ex]
    \caption{\small{An ablation study of our AOGNets using the mean error rate across 5 runs. In the first two rows, the AOGNets use full structure, and the pruned structure in the last two rows. The feature dimensions of node operations are accordingly specified to keep model sizes comparable.}
    }\label{table:ablation-cifar} 
\end{table}

\section{Conclusions and Discussions} \label{sec:conclusion}
This paper proposes grammar-guided network generators which construct compositional grammatical architectures for deep learning in an effective way. It presents deep AND-OR Grammar networks (AOGNets). The AOG comprises a phrase structure grammar and a dependency grammar. An AOGNet consists of a number of stages each of which comprises a number of AOG building blocks. Our AOG building block harnesses the best of grammar models and DNNs for deep learning. AOGNet obtains state-of-the-art performance. 
In CIFAR-10/100~\cite{CIFAR} and ImageNet-1K~\cite{ImageNet}, AOGNet obtains better performance than all state-of-the-art networks under fair comparisons. AOGNet also obtains the best model interpretability score using network dissection~\cite{netdissect2017}. AOGNet further shows better potential in adversarial defense. In MS-COCO~\cite{COCO}, AOGNet obtains better performance than the ResNet and ResNeXt backbones in Mask R-CNN~\cite{MaskRCNN}.  

\textbf{Discussions.} We hope this paper encourages further exploration in learning grammar-guided network generators. The AOG can be easily extended to adopt $k$-branch splitting rules with $k>2$. Other types of edges can also be easily introduced in the AOG such as dense lateral connections and top-down connections. Node operations can also be extended to exploit grammar-guided transformation. And, better parameter initialization methods need to be studied for the AOG structure.
  
\section*{Ackowledgement} 
This work is supported by ARO grant W911NF1810295 and DoD DURIP grant W911NF1810209. Some early experiments used the XSEDE~\cite{XSEDE} at the SDSC Comet GPU Nodes through allocation IRI180024 (XSEDE is supported by NSF grant ACI-1548562).

{\small
\bibliographystyle{ieee}
\bibliography{references}

\begin{thebibliography}{10}\itemsep=-1pt

\bibitem{BoundsOfDL}
Sanjeev Arora, Aditya Bhaskara, Rong Ge, and Tengyu Ma.
\newblock Provable bounds for learning some deep representations.
\newblock In {\em Proceedings of the 31th International Conference on Machine
  Learning, {ICML}}, pages 584--592, 2014.

\bibitem{AdversarialExampl}
Anish Athalye and Ilya Sutskever.
\newblock Synthesizing robust adversarial examples.
\newblock {\em CoRR}, abs/1707.07397, 2017.

\bibitem{DAGRNN}
Pierre Baldi and Gianluca Pollastri.
\newblock The principled design of large-scale recursive neural network
  architectures--dag-rnns and the protein structure prediction problem.
\newblock {\em Journal of Machine Learning Research}, 4:575--602, 2003.

\bibitem{netdissect2017}
David Bau, Bolei Zhou, Aditya Khosla, Aude Oliva, and Antonio Torralba.
\newblock Network dissection: Quantifying interpretability of deep visual
  representations.
\newblock In {\em Proceedings of the IEEE Conference on Computer Vision and
  Pattern Recognition}, 2017.

\bibitem{VGG}
Ken Chatfield, Karen Simonyan, Andrea Vedaldi, and Andrew Zisserman.
\newblock Return of the devil in the details: Delving deep into convolutional
  nets.
\newblock In {\em British Machine Vision Conference, {BMVC}}, 2014.

\bibitem{AOG_TaskPlanning}
Tianshui Chen, Riquan Chen, Lin Nie, Xiaonan Luo, Xiaobai Liu, and Liang Lin.
\newblock Neural task planning with {AND-OR} graph representations.
\newblock {\em {IEEE} Trans. Multimedia}, 21(4):1022--1034, 2019.

\bibitem{DPN}
Yunpeng Chen, Jianan Li, Huaxin Xiao, Xiaojie Jin, Shuicheng Yan, and Jiashi
  Feng.
\newblock Dual path networks.
\newblock {\em arXiv preprint arXiv:1707.01629}, 2017.

\bibitem{XAI}
DARPA.
\newblock Explainable artificial intelligence (xai) program,
  http://www.darpa.mil/program/ explainable-artificial-intelligence, full
  solicitation at http://www.darpa.mil/attachments/ darpa-baa-16-53.pdf.

\bibitem{NAS-survey1}
Thomas Elsken, Jan~Hendrik Metzen, and Frank Hutter.
\newblock Neural architecture search: {A} survey.
\newblock {\em CoRR}, abs/1808.05377, 2018.

\bibitem{Pff_Grammar}
Pedro~F. Felzenszwalb.
\newblock Object detection grammars.
\newblock In {\em {IEEE} International Conference on Computer Vision Workshops,
  {ICCV}}, page 691, 2011.

\bibitem{DPM}
Pedro~F. Felzenszwalb, Ross~B. Girshick, David McAllester, and Deva Ramanan.
\newblock Object detection with discriminatively trained part-based models.
\newblock {\em IEEE Trans. Pattern Anal. Mach. Intell. (PAMI)},
  32(9):1627--1645, Sept. 2010.

\bibitem{Syntactic}
King~Sun Fu and J.~E. Albus, editors.
\newblock {\em Syntactic pattern recognition : applications}.
\newblock Communication and cybernetics. Springer-Verlag, Berlin, New York,
  1977.

\bibitem{Geman_CompositionSystems}
Stuart Geman, Daniel Potter, and Zhi~Yi Chi.
\newblock Composition systems.
\newblock {\em Quarterly of Applied Mathematics}, 60(4):707--736, 2002.

\bibitem{RCN}
D. George, W. Lehrach, K. Kansky, M. L{\'a}zaro-Gredilla, C. Laan, B. Marthi,
  X. Lou, Z. Meng, Y. Liu, H. Wang, A. Lavin, and D.~S. Phoenix.
\newblock A generative vision model that trains with high data efficiency and
  breaks text-based captchas.
\newblock {\em Science}, 2017.

\bibitem{gholami2018squeezenext}
Amir Gholami, Kiseok Kwon, Bichen Wu, Zizheng Tai, Xiangyu Yue, Peter Jin,
  Sicheng Zhao, and Kurt Keutzer.
\newblock Squeezenext: Hardware-aware neural network design.
\newblock {\em arXiv preprint arXiv:1803.10615}, 2018.

\bibitem{FGSM}
Ian Goodfellow, Jonathon Shlens, and Christian Szegedy.
\newblock Explaining and harnessing adversarial examples.
\newblock In {\em ICLR}, 2015.

\bibitem{TrainImageNet1Hour}
Priya Goyal, Piotr Doll{\'{a}}r, Ross~B. Girshick, Pieter Noordhuis, Lukasz
  Wesolowski, Aapo Kyrola, Andrew Tulloch, Yangqing Jia, and Kaiming He.
\newblock Accurate, large minibatch {SGD:} training imagenet in 1 hour.
\newblock {\em CoRR}, abs/1706.02677, 2017.

\bibitem{DAGRNN1}
Alex Graves and J{\"{u}}rgen Schmidhuber.
\newblock Offline handwriting recognition with multidimensional recurrent
  neural networks.
\newblock In {\em Advances in Neural Information Processing Systems 21,
  Proceedings of the Twenty-Second Annual Conference on Neural Information
  Processing Systems, Vancouver, British Columbia, Canada, December 8-11,
  2008}, pages 545--552, 2008.

\bibitem{Grenander_PT}
Ulf Grenander and Michael Miller.
\newblock {\em Pattern Theory: From Representation to Inference}.
\newblock Oxford University Press, 2007.

\bibitem{DPRN}
Dongyoon Han, Jiwhan Kim, and Junmo Kim.
\newblock Deep pyramidal residual networks.
\newblock {\em IEEE CVPR}, 2017.

\bibitem{DependencyGrammar}
David~G. Hays.
\newblock Dependency theory: A formalism and some observations.
\newblock {\em Language}, 40(4):511--525, 1964.

\bibitem{MaskRCNN}
Kaiming He, Georgia Gkioxari, Piotr Doll{\'{a}}r, and Ross~B. Girshick.
\newblock Mask {R-CNN}.
\newblock In {\em {IEEE} International Conference on Computer Vision, {ICCV}
  2017, Venice, Italy, October 22-29, 2017}, pages 2980--2988, 2017.

\bibitem{ResidualNet}
Kaiming He, Xiangyu Zhang, Shaoqing Ren, and Jian Sun.
\newblock Deep residual learning for image recognition.
\newblock In {\em IEEE Conference on Computer Vision and Pattern Recognition
  (CVPR)}, 2016.

\bibitem{ResNetPreAct}
Kaiming He, Xiangyu Zhang, Shaoqing Ren, and Jian Sun.
\newblock Identity mappings in deep residual networks.
\newblock In {\em Computer Vision - {ECCV} 2016 - 14th European Conference,
  Amsterdam, The Netherlands, October 11-14, 2016, Proceedings, Part {IV}},
  pages 630--645, 2016.

\bibitem{Hinton}
Geoffrey Hinton.
\newblock What is wrong with convolutional neural nets?
\newblock {\em the 2017 - 2018 Machine Learning Advances and Applications
  Seminar presented by the Vector Institute at U of Toronto,
  https://www.youtube.com/watch?v=Mqt8fs6ZbHk}, August 17, 2017.

\bibitem{howard2017mobilenets}
Andrew~G Howard, Menglong Zhu, Bo Chen, Dmitry Kalenichenko, Weijun Wang,
  Tobias Weyand, Marco Andreetto, and Hartwig Adam.
\newblock Mobilenets: Efficient convolutional neural networks for mobile vision
  applications.
\newblock {\em arXiv preprint arXiv:1704.04861}, 2017.

\bibitem{SENet}
Jie Hu, Li Shen, and Gang Sun.
\newblock Squeeze-and-excitation networks.
\newblock {\em CoRR}, abs/1709.01507, 2017.

\bibitem{huang2017condensenet}
Gao Huang, Shichen Liu, Laurens van~der Maaten, and Kilian~Q Weinberger.
\newblock Condensenet: An efficient densenet using learned group convolutions.
\newblock {\em group}, 3(12):11, 2017.

\bibitem{DenseNet}
Gao Huang, Zhuang Liu, Laurens van~der Maaten, and Kilian~Q Weinberger.
\newblock Densely connected convolutional networks.
\newblock In {\em Proceedings of the IEEE Conference on Computer Vision and
  Pattern Recognition}, 2017.

\bibitem{StochasticResNet}
Gao Huang, Yu Sun, Zhuang Liu, Daniel Sedra, and Kilian~Q. Weinberger.
\newblock Deep networks with stochastic depth.
\newblock In {\em Computer Vision - {ECCV} 2016 - 14th European Conference,
  Amsterdam, The Netherlands, October 11-14, 2016, Proceedings, Part {IV}},
  pages 646--661, 2016.

\bibitem{BatchNorm}
Sergey Ioffe and Christian Szegedy.
\newblock Batch normalization: Accelerating deep network training by reducing
  internal covariate shift.
\newblock In David Blei and Francis Bach, editors, {\em Proceedings of the 32nd
  International Conference on Machine Learning (ICML-15)}, pages 448--456. JMLR
  Workshop and Conference Proceedings, 2015.

\bibitem{CIFAR}
A. Krizhevsky and G. Hinton.
\newblock Learning multiple layers of features from tiny images.
\newblock {\em Master's thesis, Department of Computer Science, University of
  Toronto}, 2009.

\bibitem{AlexNet}
Alex Krizhevsky, Ilya Sutskever, and Geoffrey~E. Hinton.
\newblock Imagenet classification with deep convolutional neural networks.
\newblock In {\em Neural Information Processing Systems (NIPS)}, pages
  1106--1114, 2012.

\bibitem{ProbabilisticProgramInduction}
Brenden~M. Lake, Ruslan Salakhutdinov, and Joshua~B. Tenenbaum.
\newblock Human-level concept learning through probabilistic program induction.
\newblock {\em Science}, 350(6266):1332--1338, 2015.

\bibitem{lake15science}
B.~M. Lake, R. Salakhutdinov, and J.~B. Tenenbaum.
\newblock Human-level concept learning through probabilistic program induction.
\newblock {\em Science}, 2015.

\bibitem{BuildMachineLikePeople}
Brenden~M. Lake, Tomer~D. Ullman, Joshua~B. Tenenbaum, and Samuel~J. Gershman.
\newblock Building machines that learn and think like people.
\newblock {\em CoRR}, abs/1604.00289, 2016.

\bibitem{FractalNet}
Gustav Larsson, Michael Maire, and Gregory Shakhnarovich.
\newblock Fractalnet: Ultra-deep neural networks without residuals.
\newblock {\em CoRR}, abs/1605.07648, 2016.

\bibitem{LeCunCNN}
Yann LeCun, Leon Bottou, Yoshua Bengio, and Patrick Haffner.
\newblock Gradient-based learning applied to document recognition.
\newblock {\em Proceedings of the IEEE}, 86(11):2278--2324, 1998.

\bibitem{AOG_Action}
Xiaodan Liang, Liang Lin, and Liangliang Cao.
\newblock Learning latent spatio-temporal compositional model for human action
  recognition.
\newblock {\em CoRR}, abs/1502.00258, 2015.

\bibitem{AOG_Shape}
Liang Lin, Xiaolong Wang, Wei Yang, and Jian{-}Huang Lai.
\newblock Discriminatively trained and-or graph models for object shape
  detection.
\newblock {\em {IEEE} Trans. Pattern Anal. Mach. Intell.}, 37(5):959--972,
  2015.

\bibitem{StochasticGrammar}
Liang Lin, Tianfu Wu, Jake Porway, and Zijian Xu.
\newblock A stochastic graph grammar for compositional object representation
  and recognition.
\newblock {\em Pattern Recognition}, 42(7):1297--1307, 2009.

\bibitem{NetInNet}
Min Lin, Qiang Chen, and Shuicheng Yan.
\newblock Network in network.
\newblock {\em CoRR}, abs/1312.4400, 2013.

\bibitem{COCO}
Tsung{-}Yi Lin, Michael Maire, Serge~J. Belongie, Lubomir~D. Bourdev, Ross~B.
  Girshick, James Hays, Pietro Perona, Deva Ramanan, Piotr Doll{\'{a}}r, and
  C.~Lawrence Zitnick.
\newblock Microsoft {COCO:} common objects in context.
\newblock {\em CoRR}, abs/1405.0312, 2014.

\bibitem{cosine_lr}
Ilya Loshchilov and Frank Hutter.
\newblock {SGDR:} stochastic gradient descent with restarts.
\newblock {\em CoRR}, abs/1608.03983, 2016.

\bibitem{maskrcnn-code}
Francisco Massa and Ross Girshick.
\newblock {maskrcnn-benchmark: Fast, modular reference implementation of
  Instance Segmentation and Object Detection algorithms in PyTorch}.
\newblock \url{https://github.com/facebookresearch/maskrcnn-benchmark}, 2018.
\newblock Accessed: [Insert date here].

\bibitem{Mumford}
David Mumford.
\newblock Grammar isn't merely part of language.
\newblock \url{http://www.dam.brown.edu/people/mumford/blog/2016/grammar.html}.

\bibitem{Mumford_PT}
D. Mumford and A. Desolneux.
\newblock {\em Pattern Theory, the Stochastic Analysis of Real World Signals}.
\newblock AKPeters/CRC Press, 2010.

\bibitem{Hourglass}
Alejandro Newell, Kaiyu Yang, and Jia Deng.
\newblock Stacked hourglass networks for human pose estimation.
\newblock {\em CoRR}, abs/1603.06937, 2016.

\bibitem{park2018bam}
Jongchan Park, Sanghyun Woo, Joon-Young Lee, and In~So Kweon.
\newblock Bam: bottleneck attention module.
\newblock {\em arXiv preprint arXiv:1807.06514}, 2018.

\bibitem{DenseNet-efficient}
Geoff Pleiss, Danlu Chen, Gao Huang, Tongcheng Li, Laurens van~der Maaten, and
  Kilian~Q. Weinberger.
\newblock Memory-efficient implementation of densenets.
\newblock {\em CoRR}, abs/1707.06990, 2017.

\bibitem{Foolbox}
Jonas Rauber, Wieland Brendel, and Matthias Bethge.
\newblock Foolbox: A python toolbox to benchmark the robustness of machine
  learning models.
\newblock {\em arXiv preprint arXiv:1707.04131}, 2017.

\bibitem{ImageNet}
Olga Russakovsky, Jia Deng, Hao Su, Jonathan Krause, Sanjeev Satheesh, Sean Ma,
  Zhiheng Huang, Andrej Karpathy, Aditya Khosla, Michael Bernstein,
  Alexander~C. Berg, and Li Fei-Fei.
\newblock {ImageNet Large Scale Visual Recognition Challenge}.
\newblock {\em Int. J. Comput. Vision (IJCV)}, 115(3):211--252, 2015.

\bibitem{capsules}
S. {Sabour}, N. {Frosst}, and G. {E Hinton}.
\newblock {Dynamic Routing Between Capsules}.
\newblock {\em ArXiv e-prints}, Oct. 2017.

\bibitem{sandler2018mobilenetv2}
Mark Sandler, Andrew Howard, Menglong Zhu, Andrey Zhmoginov, and Liang-Chieh
  Chen.
\newblock Mobilenetv2: Inverted residuals and linear bottlenecks.
\newblock In {\em Proceedings of the IEEE Conference on Computer Vision and
  Pattern Recognition}, pages 4510--4520, 2018.

\bibitem{DisAOT-CVPR}
Xi Song, Tianfu Wu, Yunde Jia, and Song{-}Chun Zhu.
\newblock Discriminatively trained and-or tree models for object detection.
\newblock In {\em Proceedings of 2013 {IEEE} Conference on Computer Vision and
  Pattern Recognition ({CVPR})}, pages 3278--3285, 2013.

\bibitem{HighwayNet}
Rupesh~Kumar Srivastava, Klaus Greff, and J{\"{u}}rgen Schmidhuber.
\newblock Highway networks.
\newblock {\em CoRR}, abs/1505.00387, 2015.

\bibitem{ProbabilisticPrograming}
Phillip~D. Summers.
\newblock A methodology for {LISP} program construction from examples.
\newblock {\em J. {ACM}}, 24(1):161--175, 1977.

\bibitem{InceptionNet}
Christian Szegedy, Sergey Ioffe, and Vincent Vanhoucke.
\newblock Inception-v4, inception-resnet and the impact of residual connections
  on learning.
\newblock {\em CoRR}, abs/1602.07261, 2016.

\bibitem{GoogLeNet}
Christian Szegedy, Wei Liu, Yangqing Jia, Pierre Sermanet, Scott~E. Reed,
  Dragomir Anguelov, Dumitru Erhan, Vincent Vanhoucke, and Andrew Rabinovich.
\newblock Going deeper with convolutions.
\newblock In {\em {IEEE} Conference on Computer Vision and Pattern Recognition,
  {CVPR} 2015, Boston, MA, USA, June 7-12, 2015}, pages 1--9, 2015.

\bibitem{YingWu_CompositionalPose}
Wei Tang, Pei Yu, and Ying Wu.
\newblock Deeply learned compositional models for human pose estimation.
\newblock In {\em {ECCV} {(3)}}, volume 11207 of {\em Lecture Notes in Computer
  Science}, pages 197--214. Springer, 2018.

\bibitem{YingWu_CompositionalModel}
Wei Tang, Pei Yu, Jiahuan Zhou, and Ying Wu.
\newblock Towards a unified compositional model for visual pattern modeling.
\newblock In {\em {ICCV}}, pages 2803--2812. {IEEE} Computer Society, 2017.

\bibitem{XSEDE}
John Towns, Timothy Cockerill, Maytal Dahan, Ian~T. Foster, Kelly~P. Gaither,
  Andrew~S. Grimshaw, Victor Hazlewood, Scott Lathrop, David Lifka, Gregory~D.
  Peterson, Ralph Roskies, J.~Ray Scott, and Nancy Wilkins{-}Diehr.
\newblock {XSEDE:} accelerating scientific discovery.
\newblock {\em Computing in Science and Engineering}, 16(5):62--74, 2014.

\bibitem{DFN}
Jingdong Wang, Zhen Wei, Ting Zhang, and Wenjun Zeng.
\newblock Deeply-fused nets.
\newblock {\em CoRR}, abs/1605.07716, 2016.

\bibitem{woo2018cbam}
Sanghyun Woo, Jongchan Park, Joon-Young Lee, and In So~Kweon.
\newblock Cbam: Convolutional block attention module.
\newblock In {\em Proceedings of the European Conference on Computer Vision
  (ECCV)}, pages 3--19, 2018.

\bibitem{TLP-PAMI}
Tianfu Wu, Yang Lu, and Song{-}Chun Zhu.
\newblock Online object tracking, learning and parsing with and-or graphs.
\newblock {\em {IEEE} Trans. Pattern Anal. Mach. Intell. ({PAMI})}, DOI:
  10.1109/TPAMI.2016.2644963, 2016.

\bibitem{ResNeXt}
Saining Xie, Ross~B. Girshick, Piotr Doll{\'{a}}r, Zhuowen Tu, and Kaiming He.
\newblock Aggregated residual transformations for deep neural networks.
\newblock {\em CoRR}, abs/1611.05431, 2016.

\bibitem{DAGCNN}
Songfan Yang and Deva Ramanan.
\newblock Multi-scale recognition with dag-cnns.
\newblock In {\em 2015 {IEEE} International Conference on Computer Vision,
  {ICCV} 2015, Santiago, Chile, December 7-13, 2015}, pages 1215--1223, 2015.

\bibitem{NAS-survey2}
Quanming Yao, Mengshuo Wang, Hugo~Jair Escalante, Isabelle Guyon, Yi{-}Qi Hu,
  Yu{-}Feng Li, Wei{-}Wei Tu, Qiang Yang, and Yang Yu.
\newblock Taking human out of learning applications: {A} survey on automated
  machine learning.
\newblock {\em CoRR}, abs/1810.13306, 2018.

\bibitem{DLA}
Fisher Yu, Dequan Wang, and Trevor Darrell.
\newblock Deep layer aggregation.
\newblock In {\em CVPR}, 2018.

\bibitem{WideResNet}
Sergey Zagoruyko and Nikos Komodakis.
\newblock Wide residual networks.
\newblock In {\em Proceedings of the British Machine Vision Conference 2016,
  {BMVC} 2016, York, UK, September 19-22, 2016}, 2016.

\bibitem{IGC1}
Ting Zhang, Guo{-}Jun Qi, Bin Xiao, and Jingdong Wang.
\newblock Interleaved group convolutions.
\newblock In {\em {IEEE} International Conference on Computer Vision, {ICCV}
  2017, Venice, Italy, October 22-29, 2017}, pages 4383--4392, 2017.

\bibitem{PolyNet}
Xingcheng Zhang, Zhizhong Li, Chen~Change Loy, and Dahua Lin.
\newblock Polynet: {A} pursuit of structural diversity in very deep networks.
\newblock {\em CoRR}, abs/1611.05725, 2016.

\bibitem{zhang2017shufflenet}
Xiangyu Zhang, Xinyu Zhou, Mengxiao Lin, and Jian Sun.
\newblock Shufflenet: An extremely efficient convolutional neural network for
  mobile devices.
\newblock {\em arXiv preprint arXiv:1707.01083}, 2017.

\bibitem{Yuille_AndOr}
Long Zhu, Yuanhao Chen, Yifei Lu, Chenxi Lin, and Alan~L. Yuille.
\newblock Max margin {AND/OR} graph learning for parsing the human body.
\newblock In {\em 2008 {IEEE} Computer Society Conference on Computer Vision
  and Pattern Recognition (CVPR)}, 2008.

\bibitem{Zhu_Grammar}
Song~Chun Zhu and David Mumford.
\newblock A stochastic grammar of images.
\newblock {\em Foundations and Trends in Computer Graphics and Vision},
  2(4):259--362, 2006.

\bibitem{zoph2017learning}
Barret Zoph, Vijay Vasudevan, Jonathon Shlens, and Quoc~V Le.
\newblock Learning transferable architectures for scalable image recognition.
\newblock {\em arXiv preprint arXiv:1707.07012}, 2(6), 2017.

\end{thebibliography}
}

\end{document}